\documentclass[conference]{IEEEtran}
\IEEEoverridecommandlockouts

\usepackage{amsmath,amssymb,bm}
\usepackage{mathtools}
\usepackage{graphicx}
\usepackage{booktabs}
\usepackage{multirow}
\usepackage{array}
\usepackage{tabularx}
\usepackage{xcolor}
\usepackage{microtype}
\usepackage{enumitem}
\usepackage{cite}
\usepackage{hyperref}
\hypersetup{
  colorlinks=false,   
  hidelinks
}
\usepackage{siunitx}
\DeclareSIUnit{\cells}{cells}

\newcommand{\R}{\mathbb{R}}
\newcommand{\bx}{\mathbf{x}}
\newcommand{\bz}{\mathbf{z}}
\newcommand{\btheta}{\bm{\theta}}
\newcommand{\bmu}{\bm{\mu}}
\newcommand{\bsigma}{\bm{\sigma}}

\title{Embedding-Based Federated Learning with Runtime Governance for Iron Deficiency Prediction\thanks{*These authors contributed equally to this work.}
\thanks{For the purpose of open access, the author has applied a Creative Commons Attribution (CC BY) license to any Author Accepted Manuscript version arising.}}

\author{\IEEEauthorblockN{Fan~Zhang\textsuperscript{*}}
\IEEEauthorblockA{\textit{Dept. of Applied Mathematics} \\
\textit{and Theoretical Physics} \\
\textit{University of Cambridge}\\
Cambridge, UK}
\and
\IEEEauthorblockN{Simon~Deltadahl\textsuperscript{*}}
\IEEEauthorblockA{\textit{Dept. of Applied Mathematics} \\
\textit{and Theoretical Physics} \\
\textit{University of Cambridge}\\
Cambridge, UK}
\and
\IEEEauthorblockN{Majid~Lotfian~Delouee}
\IEEEauthorblockA{\textit{Translational AI Laboratory,} \\
\textit{Dept. of Laboratory Medicine} \\
\textit{Amsterdam UMC}\\
Amsterdam, The Netherlands}
\and
\IEEEauthorblockN{Daniel~Kreuter}
\IEEEauthorblockA{\textit{Precision Health University} \\
\textit{Research Institute} \\
\textit{Queen Mary Univ. of London}\\
London, UK}
\and
\IEEEauthorblockN{Joseph~Taylor}
\IEEEauthorblockA{\textit{Precision Health University} \\
\textit{Research Institute} \\
\textit{Queen Mary Univ. of London}\\
London, UK}
\and
\IEEEauthorblockN{Allerdien~Visser}
\IEEEauthorblockA{\textit{Translational AI Laboratory,} \\
\textit{Dept. of Laboratory Medicine} \\
\textit{Amsterdam UMC}\\
Amsterdam, The Netherlands}
\and
\IEEEauthorblockN{BloodCounts!~Consortium}
\IEEEauthorblockA{\textit{Members listed in}\\
\textit{Consortium Members section}}
\and
\IEEEauthorblockN{James~H.~F.~Rudd}
\IEEEauthorblockA{\textit{Department of Medicine} \\
\textit{University of Cambridge} \\
\textit{Cambridge Biomedical Campus}\\
Cambridge, UK}
\and
\IEEEauthorblockN{Nicholas~S.~Gleadall}
\IEEEauthorblockA{\textit{NHS Blood and Transplant} \\
\textit{Cambridge Biomedical Campus}\\
Cambridge, UK}
\and
\IEEEauthorblockN{Suthesh~Sivapalaratnam}
\IEEEauthorblockA{\textit{Precision Health University} \\
\textit{Research Institute} \\
\textit{Queen Mary Univ. of London}\\
London, UK}
\and
\IEEEauthorblockN{Folkert~Asselbergs}
\IEEEauthorblockA{\textit{Dept. of Cardiology} \\
\textit{Amsterdam Cardiovascular Sciences} \\
\textit{Amsterdam UMC} \\
Amsterdam, The Netherlands}
\and
\IEEEauthorblockN{Martijn~C.~Schut}
\IEEEauthorblockA{\textit{Translational AI Laboratory,} \\
\textit{Dept. of Laboratory Medicine} \\
\textit{Amsterdam UMC}\\
Amsterdam, The Netherlands}
\and
\IEEEauthorblockN{Michael~Roberts}
\IEEEauthorblockA{\textit{Dept. of Applied Mathematics} \\
\textit{and Theoretical Physics} \\
\textit{University of Cambridge}\\
Cambridge, UK \\
mr808@cam.ac.uk}
}
\begin{document}
\bstctlcite{IEEEexample:BSTcontrol}
\maketitle

\begin{abstract}
Recent reviews find that the vast majority of published healthcare federated
learning (FL) studies never reach real-world deployment. We developed an
embedding-based FL pipeline for iron deficiency prediction from routine full blood
count (FBC) data and deployed it across real institutional environments at Amsterdam
University Medical Centre (AUMC) and NHS Blood and Transplant (NHSBT), two clinical environments that differ markedly in iron deficiency prevalence, ferritin distribution, and subject populations.
A frozen domain-specific haematology foundation model, DeepCBC, performs site-local
representation extraction, restricting federated training to a compact downstream
classifier and substantially reducing recurrent communication relative to
full-encoder federation. The two clinical datasets are structurally
not independent and identically distributed (non-IID), with heterogeneity arising
from distinct population differences rather than sampling artefacts. Runtime governance
is enforced by FLA\textsuperscript{3}, a healthcare-oriented FL platform providing
study-scoped execution, policy-based authorisation, and signed audit logging.
Standard sample-size-weighted aggregation (FedAvg) reduced the area under the
receiver operating characteristic curve (ROC-AUC) at both sites relative to
local-only training, as the global update was biased towards the larger AUMC
distribution. FedMAP, a personalised aggregation method, raised ROC-AUC from
0.9470 to 0.9594 at AUMC and from 0.8558 to 0.8671 at NHSBT relative to
local-only training, achieving the highest macro ROC-AUC of 0.9133 and the best
macro balanced accuracy overall. These results support personalised aggregation in
clinical federations where client sample size and task relevance diverge
substantially.
\end{abstract}


\section{Introduction}
\label{sec:intro}

Full blood count (FBC) testing is among the most frequently ordered investigations in
clinical practice. FBC panels do not measure iron stores directly, but indices such
as haemoglobin concentration (HGB), mean corpuscular volume (MCV), mean corpuscular haemoglobin (MCH),
and red cell distribution width (RDW) carry information relevant to iron deficiency. Prior studies show
that machine learning models built from routine laboratory data can predict low
ferritin (a proxy for low body iron stores, and the principal test used for iron deficiency diagnosis) with useful
discriminatory power~\cite{thompson1988rdw,sarma1990redcell,kurstjens2022ferritin,Kreuter2025.06.18.25329494}. The
central obstacle is not model design alone. Access to sufficiently broad data is
equally limiting: a single institution may capture too narrow a clinical
population~\cite{sheller2020federated}, and centralisation is constrained by privacy law, institutional
governance, and operational trust boundaries.

Federated learning (FL) addresses part of this problem by training models across institutions without
pooling raw patient
data~\cite{mcmahan2017communication,rieke2020future,sheller2020federated}. In
healthcare, the challenge extends beyond decentralised optimisation. Recent review
work argues that many published healthcare FL studies remain unsuitable for clinical
use citing methodological weaknesses across bias, privacy,
generalisation, communication, and governance compliance~\cite{LI2025103497, zhangRecentMethodologicalAdvances2024b, zhang2026fla3}. A complementary
systematic review found that only 5.2\% of healthcare FL studies reported real-life
application~\cite{TEO2024101419}. Genuine clinical deployment remains rare, which
motivates empirical reports from operational settings. Sites also differ in
population, indication for testing, laboratory workflow, governance requirements, and
network posture, so the problem of not independently and identically distributed data (non-IID) is often structural rather than incidental.

We report a cross-institutional study across two BloodCounts! consortium sites, AUMC and NHSBT.
The system uses the pre-trained domain-specific FBC foundation model DeepCBC \cite{Kreuter2025.06.18.25329494} for site-local
representation extraction and restricts federated training to a compact downstream
classifier. We compare local-only training with
FedAvg~\cite{mcmahan2017communication}, FedProx~\cite{li2020fedprox}, and
FedMAP~\cite{zhang2025fedmappersonalisedfederatedlearning}, a personalised aggregation method developed for
heterogeneous healthcare federations. The deployment runs on
FLA\textsuperscript{3}~\cite{zhang2026fla3}, which provides study-scoped
execution, runtime policy enforcement, and auditable logging.

This paper contributes:
\begin{enumerate}[leftmargin=*,itemsep=2pt]
  \item A practical two-stage design for healthcare FL in which a frozen haematology
    foundation model handles site-local representation extraction and only a compact
    downstream classifier is trained federatively.
  \item A characterisation of a strongly heterogeneous clinical federation in
    which sites differ substantially in prevalence, ferritin distribution, and
    effective positive-class volume, arising from distinct clinical workflows.
  \item An empirical demonstration that sample-size-weighted aggregation degrades
  performance at both sites when prevalence and clinical purpose diverge, and that
  personalised aggregation recovers and exceeds local-only ROC-AUC at both sites
  in this two-site deployment.
  \item An operational deployment demonstrating how runtime governance controls, including study scoping,
policy-based authorisation, and signed audit logging, are integrated into a
healthcare FL system.
\end{enumerate}

\section{Related Work}
\label{sec:related}

Federated learning has been applied to medical imaging, electronic health records,
and multi-centre clinical prediction, with recurrent emphasis on data heterogeneity,
generalisation, and
deployability~\cite{rieke2020future,sheller2020federated,vaid2021federated,warnat2021swarm}.
FedProx introduced a proximal term to stabilise local optimisation under
heterogeneous client distributions~\cite{li2020fedprox}. FedMAP extended this line
of work through personalised aggregation that accounts for client relevance under
heterogeneous clinical data rather than relying on dataset size
alone~\cite{zhang2025fedmappersonalisedfederatedlearning}.

Representation transfer is a complementary design consideration. A domain-specific
model can be pre-trained once, after which only a lightweight downstream head is
adapted, reducing communication volume and simplifying deployment in restrictive
institutional environments~\cite{pmlr-v232-malaviya23a}. The present work instantiates this pattern in
haematology, coupling a pre-trained FBC representation model with a federated
runtime that enforces study scope, authorisation, and auditability.

\section{Datasets and Clinical Heterogeneity}
\label{sec:data}

We study two clinically distinct cohorts drawn from BloodCounts! consortium sites.

\paragraph{AUMC cohort.}
AUMC contributes a hospital-based cohort in which ferritin testing is performed for
the diagnosis of iron deficiency, monitoring or treatment response evaluation,
exclusion of overload, or broader inpatient and outpatient workup. The cohort was
\SI{56.3}{\percent} male and \SI{43.7}{\percent} female, with median ages of
\SI{63.0}{years} and \SI{58.0}{years}, respectively. The median white blood cell
count was \SI{6.71e9}{\cells\per\litre}, with \SI{74.8}{\percent} of records below
\SI{10e9}{\cells\per\litre}. A reactive testing approach biases towards more
severe cases. Iron deficiency prevalence is lower and replete ferritin values are
comparatively high, reflecting higher levels of inflammation within the cohort.
The site is informative for evaluating specificity and generalisation in a
clinically complex setting, but it contributes relatively few positive cases.

\paragraph{NHSBT cohort.}
We use data from the INTERVAL randomised controlled trial~\cite{Di_Angelantonio_2017} assessing the safety of different blood donation frequencies (8, 10, 12 weeks for males; 12, 14, 16 weeks for females) over a 24-month period, with a subgroup also monitored up to 48 months~\cite{kaptoge_longer-term_2019}. 
The cohort is approximately balanced by sex (\SI{49.7}{\percent} male, median age 46.2 years; \SI{50.3}{\percent} female, median age 40.8 years).
This cohort is representative of a healthy population, where iron deficiency is the primary cause of anaemia. Unlike AUMC's hospital population, the inflammation burden in this population was low, with a median 
white blood cell count (WBC) of \SI{6.29e9}{\cells\per\litre} (\SI{95}{\percent} of participants with WBC under \SI{10e9}{\cells\per\litre}). Being a blood donor population, the iron deficiency prevalence was high compared to the general population at around \SI{19}{\percent}. A universal ferritin testing approach and exclusion of cases with established anaemia biases this cohort towards subclinical cases of iron deficiency, which are associated with milder changes in FBC indices and are thus difficult to detect.

\begin{table}[t]
\centering
\caption{Dataset statistics per site and split. Pos.\% is the fraction of
  samples with ferritin $<\SI{15}{\micro\gram\per\litre}$. Replete ferritin is reported for the
  training split only.}
\label{tab:data}
\small
\setlength{\tabcolsep}{3pt}
\begin{tabular}{llrrrr}
\toprule
\textbf{Site} & \textbf{Split} & \textbf{N} & \textbf{N$_+$} & \textbf{N$_-$} & \textbf{Pos.\%} \\
\midrule
\multirow{3}{*}{AUMC}
  & Train & 44{,}032 & 1{,}219 & 42{,}813 & 2.8 \\
  & Val   & 73{,}344 & 1{,}592 & 71{,}752 & 2.2 \\
  & Test  & 100{,}096 & 1{,}650 & 98{,}446 & 1.6 \\
\addlinespace
\multirow{3}{*}{NHSBT}
  & Train & 30{,}997 & 6{,}033 & 24{,}964 & 19.5 \\
  & Val   & 7{,}278  & 1{,}316 & 5{,}962  & 18.1 \\
  & Test  & 9{,}041  & 1{,}744 & 7{,}297  & 19.3 \\
\bottomrule
\multicolumn{6}{p{0.95\columnwidth}}{\small Median iron-replete ferritin, training split:
  AUMC \SI{602}{\nano\gram\per\milli\litre} (IQR \SIrange{144}{1505}{\nano\gram\per\milli\litre});
  NHSBT \SI{39}{\nano\gram\per\milli\litre} (IQR \SIrange{25}{63}{\nano\gram\per\milli\litre}).}
\end{tabular}
\end{table}

\begin{figure}[t]
  \centering
  \includegraphics[width=\columnwidth]{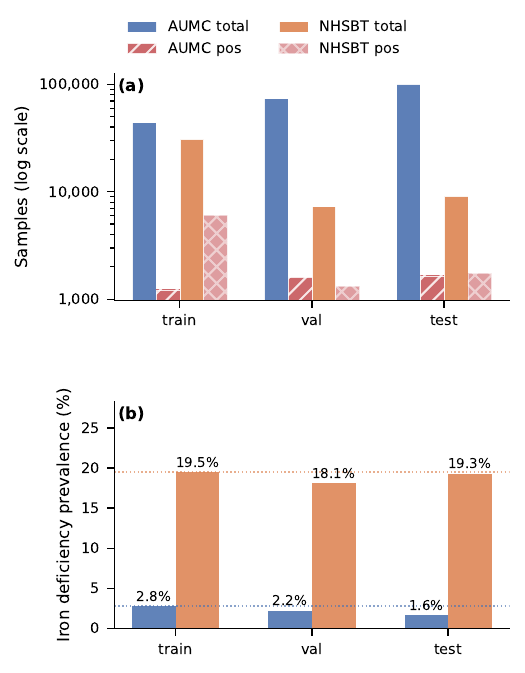}
  \caption{Cohort composition at AUMC and NHSBT. (a) Total and iron-deficient sample
    counts by split, shown on a log scale. AUMC contributes larger total sample
    volume, whereas NHSBT contributes a higher relative burden of iron-deficient
    cases. (b) Iron deficiency prevalence by split, showing a persistent prevalence
    gap between AUMC and NHSBT across training, validation, and test sets.}
  \label{fig:cohort_heterogeneity}
\end{figure}

\begin{figure}[t]
  \centering
  \includegraphics[width=\columnwidth]{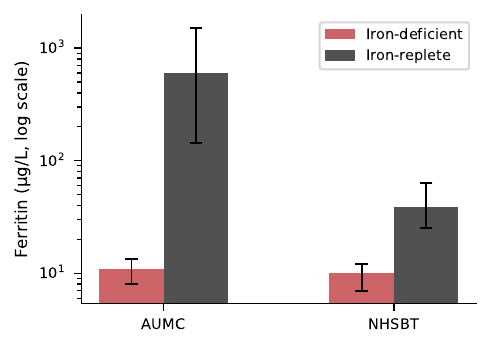}
  \caption{Training-set ferritin distributions at AUMC and NHSBT for iron-deficient
    and iron-replete groups, shown as median with IQR on a log scale. The replete
    ferritin distribution differs markedly across sites, with substantially higher
    values at AUMC.}
  \label{fig:ferritin_distribution}
\end{figure}

\begin{figure}[t]
  \centering
  \includegraphics[width=\columnwidth]{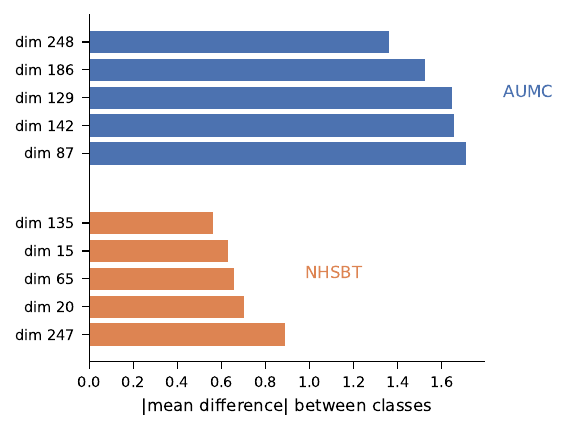}
  \caption{Top five discriminative embedding dimensions at AUMC and NHSBT, ranked
    by the absolute difference in class-conditional mean activation between
    iron-deficient and iron-replete samples.}
  \label{fig:embedding_heterogeneity}
\end{figure}

Table~\ref{tab:data} summarises the split statistics. Differences in prevalence, ferritin distribution, and effective positive-class
volume reflect genuine biological and clinical workflow differences between
the two populations.
Figs.~\ref{fig:cohort_heterogeneity} and~\ref{fig:ferritin_distribution} illustrate
the prevalence gap and the marked divergence in iron-replete ferritin values.
Figure~\ref{fig:embedding_heterogeneity} shows, for each site, the five embedding dimensions with the largest absolute difference in mean activation between iron-deficient and iron-replete samples. The ranked lists share no overlap. This divergence likely reflects the differing
biological character of the two populations: AUMC's hospital cohort carries a
higher inflammation burden, which shifts FBC indices differently than the
low-inflammation donor population at NHSBT, resulting in distinct discriminative
directions in the embedding space.

\paragraph{Label and feature heterogeneity.}
Prevalence differs by nearly an order of magnitude in the training split, from
19.5\% at NHSBT to 2.8\% at AUMC, and the gap persists across validation and test
data. Feature-level heterogeneity is also evident in the embedding space. Together,
these observations support treating the federation as structurally non-IID, with
heterogeneity driven by clinical workflow and population differences rather than
ordinary sampling noise.

\section{Methods}
\label{sec:methods}

\subsection{Task Formulation}

Each site $k$ holds a private dataset
$\mathcal{D}_k=\{(\bx_i^{(k)}, y_i^{(k)})\}_{i=1}^{N_k}$, where $\bx_i^{(k)}$ is
the patient representation derived from local FBC data and
\[
  y_i^{(k)}=\mathbf{1}[\mathrm{ferritin}_i^{(k)}<\SI{15}{\micro\gram\per\litre}]
\]
indicates iron deficiency in line with world health organisation guidelines \cite{worldhealthorganisationWHOGuidelineUse2020}. The objective is to
learn a classifier $f_{\btheta}:\R^{d}\to[0,1]$ without transferring raw patient
records between institutions, where the embedding dimensionality $d$ is determined
by the site-local feature extractor (Section~\ref{sec:embeddings}).

\subsection{Embedding Extraction}
\label{sec:embeddings}

The first stage uses DeepCBC, a foundation model trained previously on
large-scale flow cytometry and impedance data underlying the FBC (\textit{raw} FBC data) \cite{Kreuter2025.06.18.25329494}. At deployment time only the frozen encoder is used. Given
a standardised local FBC input $\bx_i$, the encoder defines a latent posterior
\[
  q_{\phi}(\bz \mid \bx_i)=\mathcal{N}\!\big(\bmu_{\phi}(\bx_i),
  \mathrm{diag}(\bsigma^2_{\phi}(\bx_i))\big).
\]
For downstream classification we use the posterior mean as a deterministic
embedding,
\[
  \bz_i=\bmu_{\phi}(\bx_i)\in\R^{256}.
\]
This choice removes stochasticity at inference time and provides a fixed-length
representation for each patient. Raw FBC inputs and patient-level embeddings remain inside the institutional
environment, the
foundation model is distributed once as a local artefact, and recurrent federated
communication is limited to classifier parameters.

This two-stage decomposition is motivated by both optimisation and deployment
constraints. Adaptation is concentrated at the downstream decision boundary rather
than requiring joint representation learning across highly heterogeneous cohorts.
Communication overhead is reduced relative to a federation that repeatedly
synchronises a larger encoder. In this deployment, recurrent communication is
limited to the downstream multilayer perceptron (MLP) classifier parameters, whereas synchronising the
full DeepCBC encoder at the same cadence would require transmitting roughly three orders of magnitude more parameters per round.

\subsection{Federated Classifier and Aggregation}
\label{sec:fl}

The downstream classifier is a two-hidden-layer MLP ($256 \rightarrow 128 \rightarrow 64 \rightarrow 1$) with     
  batch normalisation, ReLU activations, and dropout ($p = 0.3$), totalling 41{,}601 trainable parameters. Local-only training serves as the non-federated baseline at each site.

Three federated aggregation strategies are compared:
\begin{itemize}[leftmargin=*,itemsep=1pt]
  \item \textbf{FedAvg}: the standard sample-size-weighted averaging
    baseline~\cite{mcmahan2017communication}.
  \item \textbf{FedProx}: local optimisation with a proximal penalty
    $\frac{\mu}{2}\|\btheta-\btheta^{(t)}\|_2^2$ to reduce client drift under
    heterogeneity~\cite{li2020fedprox}.
  \item \textbf{FedMAP}: a personalised aggregation framework in which local
    optimisation is cast as maximum a posteriori (MAP) estimation under a learned
    regulariser $R(\theta;\mu_g,\psi)$, and server aggregation uses
    posterior-informed weights rather than sample count
    alone~\cite{zhang2025fedmappersonalisedfederatedlearning}.
\end{itemize}

FedMAP is summarised here; the full derivation is given in the original paper.
FedMAP learns a global regulariser of the form
\[
  R(\theta;\theta_g,\psi)=f_{\psi}(\theta,\theta_g)+\alpha \lVert \theta-\theta_g \rVert^2
  + \epsilon (\lVert \theta \rVert^2 + \lVert \theta_g \rVert^2),
\]
where $f_{\psi}$ is an input-convex neural network (ICNN), $\theta_g$ denotes the
global model parameters, and $\psi$ parameterises the learned regularisation
landscape. At each round, site $k$ updates its local classifier
by MAP optimisation and returns both parameters and an aggregation weight, presented
in summary form from:
\[
 \omega_k^{(t)} \propto P(\mathcal{D}_k \mid \theta_k^{(t+1)})
  \exp\!\big(-R(\theta_k^{(t+1)};\theta_g^{(t)},\psi^{(t)})\big),
\]
where $\mathcal{D}_k$ is the local dataset at site $k$. This weight favours local
models that fit their data well and remain close to the globally learned
prior. The server then updates the global model by weighted
averaging and refines $\psi$ through gradient steps on the regulariser objective.

All federated methods used 50 communication rounds, batch size 256, learning rate
$10^{-3}$, 10 local epochs per round, and early stopping with patience 3.
Thresholds were tuned on the validation split (Table~\ref{tab:data}). FedProx used proximal coefficient
$\mu_p=0.05$. FedMAP used an ICNN learning rate of $10^{-5}$ and 3 ICNN steps per
round.

\section{FLA\textsuperscript{3} Platform and Runtime Governance}
\label{sec:fla3}

Governance in this deployment is enforced at runtime rather than recorded as a
static agreement outside the system. The federation therefore runs on
FLA\textsuperscript{3}~\cite{zhang2026fla3}, a healthcare-oriented federated
learning platform built around Flower-based~\cite{beutel2022flowerfriendlyfederatedlearning} orchestration with explicit authorisation
and auditing.

Each study is scoped as a distinct federation with explicit participant roles. The
orchestration layer separates central coordination from site-local execution,
corresponding broadly to Flower SuperLink and SuperNode responsibilities in the
platform design. Authorisation is policy-driven and compliant with the eXtensible Access Control
Markup Language (XACML), with deny-by-default semantics: an action is permitted
only if an explicit rule grants it. Key execution events are recorded in an
append-only signed audit trail, allowing institutions to reconstruct who requested a
training action, under which policy, and at what time. In this deployment, a
site-local node attempting to push parameters outside its assigned aggregation
window is denied by the XACML policy engine before
any network transmission occurs; the denial is recorded in the audit log with the
policy rule identifier and timestamp.

Participating organisations in multi-institutional healthcare FL often require
assurance beyond local data retention. Role-bound execution, temporal scoping,
policy compliance, and post hoc accountability are distinct requirements that
governance documentation alone cannot satisfy. FLA\textsuperscript{3} provides
those controls as constituents of the training runtime rather than as an external
wrapper.

\section{Experiments, Results and Discussion}
\label{sec:eval}

Each site maintained separate train, validation, and test splits. The primary metric
was ROC--AUC, reported separately for AUMC and NHSBT and summarised as a macro
average across sites. Balanced accuracy was also reported as class distributions
differed sharply across sites. Threshold-dependent metrics used thresholds selected
on the validation split. Macro ROC--AUC and macro balanced accuracy were computed as the mean of the
two per-site values. Confidence intervals are reported to summarise uncertainty.

The primary interpretive question concerns whether federation improves performance
at either site without material harm to the other, and whether personalised
aggregation reduces the AUMC--NHSBT performance gap.

\begin{table*}[t]
\centering
\caption{Test-set performance by site. FedMAP achieves the highest ROC--AUC at both
  sites and the highest macro balanced accuracy overall. CI: confidence interval.}
\label{tab:main}
\small
\setlength{\tabcolsep}{4pt}
\begin{tabular}{lccccc}
\toprule
\textbf{Method} & \textbf{Site} & \textbf{ROC--AUC} & \textbf{95\% CI}
  & \textbf{Bal.Acc} & \textbf{95\% CI} \\
\midrule
Local-only & AUMC  & 0.9470 & [0.9394, 0.9545] & \textbf{0.8964} & \textbf{[0.8873, 0.9046]}\\
Local-only & NHSBT & 0.8558 & [0.8442, 0.8673] & 0.7344 & [0.7217, 0.7462] \\
FedAvg     & AUMC  & 0.9380 & [0.9339, 0.9422] & 0.8436 & [0.8320, 0.8546] \\
FedAvg     & NHSBT & 0.8432 & [0.8345, 0.8520] & 0.7719 & [0.7570, 0.7860] \\
FedProx    & AUMC  & 0.9446 & [0.9401, 0.9492] & 0.8539 & [0.8426, 0.8645] \\
FedProx    & NHSBT & 0.8487 & [0.8447, 0.8527] & 0.7774 & [0.7627, 0.7912] \\
FedMAP     & AUMC  & \textbf{0.9594} & \textbf{[0.9590, 0.9598]}
           & 0.8899 & [0.8811, 0.8978] \\
FedMAP     & NHSBT & \textbf{0.8671} & \textbf{[0.8652, 0.8690]}
           & \textbf{0.7879} & \textbf{[0.7732, 0.8018]} \\
\midrule
Local-only & Macro & 0.9014 & -- & 0.8154 & -- \\
FedAvg     & Macro & 0.8906 & -- & 0.8078 & -- \\
FedProx    & Macro & 0.8967 & -- & 0.8157 & -- \\
FedMAP     & Macro & \textbf{0.9133} & -- & \textbf{0.8389} & -- \\
\bottomrule
\end{tabular}
\end{table*}

Table~\ref{tab:main} reports the main test-set results. FedAvg reduced ROC--AUC
relative to local-only training at both AUMC and NHSBT. AUMC contributes roughly
1.4 times the total training samples of NHSBT, so sample-size-weighted averaging
biases the global update towards the AUMC distribution. Given the marked differences
in prevalence and embedding-level class structure between the two cohorts, this bias
degrades the decision boundary at both sites rather than improving either. FedProx recovered part of this loss but did not achieve the best overall performance; its proximal penalty stabilises local optimisation around a shared global model, but aggregation remains insensitive to client task relevance. It therefore addresses client drift without resolving the underlying mismatch between sample-size weighting and structural clinical heterogeneity. FedMAP
achieved the highest ROC--AUC at both sites, improving from 0.9470 to 0.9594 at
AUMC and from 0.8558 to 0.8671 at NHSBT compared to local-only training, with the highest macro ROC--AUC of
0.9133. The discrepancy in performance in terms of ROC-AUC is likely to reflect difference in iron deficiency case severity between sites, resulting from the different ferritin testing approach (universal vs reactive) described above.

Balanced accuracy showed a similar but not identical pattern. At NHSBT, all
federated methods improved on the local-only baseline, with FedMAP reaching 0.7879.
At AUMC, local-only training remained highest at 0.8964; FedMAP was close at 0.8899
and outperformed FedAvg and FedProx. The small balanced accuracy gap at AUMC
(0.0065) likely reflects threshold sensitivity under the site's low positive
prevalence: personalised aggregation shifts the global prior in a direction that
improves discrimination (ROC--AUC) without a corresponding shift in the optimal
classification threshold, so threshold-dependent metrics do not fully capture the
gain. FedMAP achieved the highest macro balanced accuracy of 0.8389. Of note, in these
experiments we do not allow for threshold re-optimisation post-aggregation by site
and use the World Health Organisation recommended threshold of
\SI{15}{\micro\gram\per\litre}; however, due to the higher levels of inflammation
in the AUMC hospital cohort we expect an iron-status independent right shift in
ferritin values and a corresponding loss of test sensitivity. This effect is
demonstrated in the higher median ferritin in iron repletion seen in the AUMC
cohort (Figure~\ref{fig:ferritin_distribution}). In clinical practice a higher
threshold of \SI{30}{\micro\gram\per\litre} is frequently employed as a crude
form of threshold re-optimisation to adjust for this problem~\cite{Fletcher2022}.

These findings support the view that personalised aggregation is beneficial when
client sample size and task relevance diverge across sites.

\section{Lessons Learned and Practical Implications}
\label{sec:lessons}

\paragraph*{Lesson 1: structural heterogeneity should be treated as a clinical
  property, not only a statistical nuisance}
We applied the same ferritin threshold to define iron deficiency for NHSBT and AUMC, but
their cohorts arise from different clinical workflows: NHSBT reflects donor
screening, whereas AUMC reflects hospital-based testing and monitoring. The
resulting prevalence and ferritin-distribution differences indicate that the sites
do not represent the same task distribution. For healthcare FL studies, reporting
class balance alone is insufficient; the clinical origin of the label must also be
described.

\paragraph*{Lesson 2: embedding-based federation offers a practical deployment
  pattern for hospitals}
A frozen domain-specific foundation model can be used for local representation
extraction, reducing the federated task to a compact downstream classifier. This
lowers communication burden and simplifies deployment inside institutional
environments. It retains the benefits of domain-specific pretraining without
repeated synchronisation of a larger encoder. In this deployment, recurrent
communication is limited to the downstream MLP parameters, whereas synchronising
the full DeepCBC encoder at the same cadence would require transmitting roughly
three orders of magnitude more parameters per round. A further practical
advantage is data efficiency: individual clinical sites may hold insufficient
data to train a large representation model from scratch, whereas a frozen
pretrained encoder transfers domain knowledge that would otherwise require
far larger local datasets to acquire. In hospital settings where both
infrastructure constraints and limited local data volumes make full-model
federation difficult, separating representation learning from federated task
adaptation is a practical design choice.

\paragraph*{Lesson 3: personalised aggregation may help when client size and client
  relevance diverge}
Larger local sample volume does not necessarily imply greater relevance to the
shared task. The two cohorts differ in clinical purpose, case mix, and
representation-level class structure, and FedMAP outperformed size-weighted
baselines across the reported metrics. Healthcare FL evaluations should consider
whether aggregation reflects task alignment rather than assuming that larger sites
should dominate the global update.

\paragraph*{Lesson 4: governance should be operationalised at runtime}
Multi-institutional healthcare federation requires more than local data retention.
Without runtime enforcement, a site-local node can transmit parameters outside
its assigned aggregation window, exfiltrate embeddings, or participate beyond
its authorised study scope, flaws that static data-sharing agreements cannot
prevent after the fact. In this deployment, such violations are denied by the
XACML policy engine before any network transmission occurs and recorded in the
audit log. This runtime enforcement was a practical prerequisite for executing
the study across international borders: AUMC and NHSBT operate under distinct
regulatory frameworks (Dutch and UK clinical governance respectively), and
neither institution could have participated without verifiable, post hoc
accountable controls over how their data and model parameters were used.
Governance documentation alone cannot provide that assurance; it must be
built into the execution layer.

\paragraph*{Limitations}
Privacy attack analyses, including membership inference and gradient inversion, were
not conducted in this deployment phase and constitute an important direction for
future work. The embedding-only communication pattern reduces the gradient inversion
attack surface relative to raw-feature federation, but does not eliminate privacy
risk entirely. Runtime overhead for the personalised aggregation procedure was not
quantified; future work will profile wall-clock training time per round,
including a comparison against the cost of synchronising the full DeepCBC
encoder to characterise the communication and compute savings of the
embedding-based design.
The embedding-space heterogeneity comparison is qualitative, and formal paired
significance testing between methods was outside the scope. As evidence derives from a single two-site deployment, broader validation
across larger consortium sites would strengthen the generalisability of
these findings.
Calibration was not evaluated in this deployment phase. This is a limitation because clinical prediction models require reliable probability estimates in addition to discrimination; future deployment should include site-specific calibration assessment and recalibration where necessary before clinical use.
Because DeepCBC is frozen, a maintained deployment would require monitoring of embedding distributions for temporal drift in laboratory practice, analyser hardware, preprocessing, or patient case mix.
Future work should examine whether learned FedMAP aggregation weights can serve as quantitative indicators of inter-site heterogeneity.

\section{Ethics and Responsible Research}
\label{sec:ethics}

This study uses clinical laboratory data in a privacy-constrained setting. Raw
patient data were not centralised for model training; raw FBC measurements and
patient-level embeddings remained within each institution. Only downstream
classifier parameters were exchanged during federation. The INTERVAL trial is compiled into the Blood Donors Studies BioResource (BDSB) with Research Ethics Committee (REC) reference 20/EE/0115. The Medical Ethics Review Committee of Amsterdam University Medical Centres (registered with the US Office for Human Research Protections as IRB00013752; FWA number FWA00032965) reviewed the study protocol and determined that the Medical Research Involving Human Subjects Act (WMO) does not apply to this research; therefore, formal ethical approval and informed consent were waived.
\section{Conclusion}
\label{sec:conclusion}

We present an operational FL deployment for iron deficiency prediction from FBC
foundation-model embeddings across AUMC and NHSBT. The study brings together three
elements that are often treated separately in the literature: domain-specific
representation transfer, structural healthcare heterogeneity, and runtime governance
controls. Standard sample-size-weighted federation did not improve discrimination
across sites. FedMAP, a personalised FL algorithm, achieved the strongest overall performance, with the highest
ROC--AUC at both AUMC and NHSBT and the best macro balanced accuracy. These findings, drawn from a two-site deployment, support the case for personalised aggregation in operational clinical federations with substantial structural heterogeneity.

\section*{Acknowledgment}
F.~Zhang, S.~Deltadahl, D.~Kreuter, J.~Taylor, N.~S.~Gleadall, S.~Sivapalaratnam, and M.~Roberts have received support from the Trinity Challenge grant awarded to establish the BloodCounts! consortium, along with NIHR UCLH Biomedical Research Centre, the NIHR Cambridge Biomedical Research Centre, National Health Service Blood and Transplant (NHSBT) and the Medical Research Council. D.~Kreuter and J.~Taylor receive support from MRC GAP Fund (UKRI/814). N.~S.~Gleadall has been supported by NHSBT grant 1701-GEN. M.~Roberts is additionally supported by the British Heart Foundation (TA/F/20/210001). M.~Lotfian Delouee acknowledges support from the LabGPT project, funded by Amsterdam UMC Innovation Funding. \\
Participants in the INTERVAL randomised controlled trial were recruited with the active collaboration of NHS Blood and Transplant England (\url{www.nhsbt.nhs.uk}), which has supported field work and other elements of the trial. DNA extraction and genotyping were co-funded by the National Institute for Health and Care Research (NIHR), the NIHR BioResource (\url{http://bioresource.nihr.ac.uk}) and the NIHR Cambridge Biomedical Research Centre (BRC-1215-20014)$^\dagger$. The academic coordinating centre for INTERVAL was supported by core funding from the: NIHR Blood and Transplant Research Unit (BTRU) in Donor Health and Genomics (NIHR BTRU-2014-10024), NIHR BTRU in Donor Health and Behaviour (NIHR203337), UK Medical Research Council (MR/L003120/1), British Heart Foundation (SP/09/002; RG/13/13/30194; RG/18/13/33946), NIHR Cambridge BRC (BRC-1215-20014; NIHR203312)$^\dagger$, and by Health Data Research UK, which is funded by the UK Medical Research Council, Engineering and Physical Sciences Research Council, Economic and Social Research Council, Department of Health and Social Care (England), Chief Scientist Office of the Scottish Government Health and Social Care Directorates, Health and Social Care Research and Development Division (Welsh Government), Public Health Agency (Northern Ireland), British Heart Foundation and Wellcome. A complete list of the investigators and contributors to the INTERVAL trial is provided in~\cite{Di_Angelantonio_2017}. The academic coordinating centre would like to thank blood donor centre staff and blood donors for participating in the INTERVAL trial.

\noindent$^\dagger$\textit{The views expressed are those of the authors and not necessarily those of the NIHR or the Department of Health and Social Care.}

\section*{Conflicts of Interest}
The authors declare no conflicts of interest relevant to this work.

\section*{Code Availability}
The FedMAP personalised aggregation framework used in this work is publicly
available at \url{https://github.com/CambridgeCIA/FedMAP}~\cite{zhang2025fedmappersonalisedfederatedlearning}.
The FLA\textsuperscript{3} runtime governance platform is publicly available at
\url{https://github.com/bloodcounts/FLAAA}~\cite{zhang2026fla3}.

\section*{BloodCounts! Consortium Members}

\noindent
Martijn Schut$^{1}$, Folkert Asselbergs$^{1}$, Sujoy Kar$^{2}$,
Suthesh Sivapalaratnam$^{3}$, Sophie Williams$^{3}$, Mickey Koh$^{4}$,
Yvonne Henskens$^{5}$, Norbert C.J. de Wit$^{5}$,
Umberto D'Alessandro$^{6}$, Bubacarr Bah$^{6}$, Ousman Secka$^{6}$,
Parashkev Nachev$^{7}$, Rajeev Gupta$^{7}$, Sara Trompeter$^{7}$,
Nancy Boeckx$^{8}$, Christine van Laer$^{8}$,
Gordon A. Awandare$^{9}$, Kwabena Sarpong$^{9}$,
Lucas Amenga-Etego$^{9}$, Mathie Leers$^{10}$,
Mirelle Huijskens$^{10}$, Samuel McDermott$^{11}$,
Willem H. Ouwehand$^{12}$, James Rudd$^{13}$,
Carola-Bibiane Sch{\"o}nlieb$^{11}$,
Nicholas Gleadall$^{12,14,15}$, and Michael Roberts$^{11,13}$.

\vspace{1em}

\noindent
$^{1}$ Amsterdam University Medical Centre, Amsterdam, Netherlands.\\
$^{2}$ Apollo Hospitals, Chennai, India.\\
$^{3}$ Barts Health NHS Trust, London, United Kingdom.\\
$^{4}$ Health Services Authority, Singapore.\\
$^{5}$ Maastricht University Medical Centre, Maastricht, Netherlands.\\
$^{6}$ MRC The Gambia Unit, Banjul, The Gambia.\\
$^{7}$ University College London Hospitals, London, United Kingdom.\\
$^{8}$ University Hospitals Leuven, Leuven, Belgium.\\
$^{9}$ West African Centre for Cell Biology of Infectious Pathogens, Accra, Ghana.\\
$^{10}$ Zuyderland Medical Center, Zuyderland, Netherlands.\\
$^{11}$ Department of Applied Mathematics and Theoretical Physics, University of Cambridge, UK.\\
$^{12}$ NHS Blood and Transplant, Cambridge, UK.\\
$^{13}$ Department of Medicine, University of Cambridge, UK.\\
$^{14}$ Victor Phillip Dahdaleh Heart and Lung Research Institute, University of Cambridge, UK.\\
$^{15}$ Department of Haematology, University of Cambridge, UK.

\bibliographystyle{IEEEtran}
\bibliography{references}

\end{document}